# Semi-Synthetic Parallel Data for Translation Quality Estimation: A Case Study of Dataset Building for an Under-Resourced Language Pair

Assaf Siani, Anna Kernerman, Ilan Kernerman

{assaf, anna, ilan}@lexicala.com

**Abstract**

Quality estimation (QE) plays a crucial role in machine translation (MT) workflows, as it serves to evaluate generated outputs that have no reference translations and to determine whether human post-editing or full retranslation is necessary. Yet, developing highly accurate, adaptable and reliable QE systems for under-resourced language pairs remains largely unsolved, due mainly to limited parallel corpora and to diverse language-dependent factors, such as with morphosyntactically complex languages. This study presents a semi-synthetic parallel dataset for English-to-Hebrew QE, generated by creating English sentences based on examples of usage that illustrate typical linguistic patterns, translating them to Hebrew using multiple MT engines, and filtering outputs via BLEU-based selection. Each translated segment was manually evaluated and scored by a linguist, and we also incorporated professionally translated English-Hebrew segments from our own resources, which were assigned the highest quality score. Controlled translation errors were introduced to address linguistic challenges, particularly regarding gender and number agreement, and we trained neural QE models, including BERT and XLM-R, on this dataset to assess sentence-level MT quality. Our findings highlight the impact of dataset size, distributed balance, and error distribution on model performance. We will describe the challenges, methodology and results of our experiments, and specify future directions aimed at improving QE performance. This research contributes to advancing QE models for under-resourced language pairs, including morphology-rich languages.

## 1. Introduction

Quality Estimation (QE) is vital for bridging the gap between human and machine translation (MT), as it assesses the quality of machine-generated translations without requiring reference translations (Specia et al. 2018). Despite its importance in various applications, building highly accurate, adaptable and reliable QE systems remains a challenge, especially for under-resourced language pairs. These systems play a crucial role in optimizing translation workflows by selecting the best translation among multiple MT engines and providing users with an indication of the reliability of automated translations. Additionally, QE aids in determining whether human post-editing or full retranslation is necessary (Kepler et al. 2019).

Translation quality can be evaluated at different levels, including document, sentence, and word/phrase levels (Ive et al. 2018). This study focuses on sentence-level QE for English to Hebrew translation. Due to Hebrew's complex morphosyntactic structure and the frequent inconsistencies in MT outputs, developing a robust QE system for this language pair presents unique challenges.

To address these challenges, we constructed a semi-synthetic parallel dataset designed to enhance the reliability of QE models. The dataset was built by generating synthetic English sentences based on reference examples and translating them into Hebrew using multiple MT engines. The translations were then systematically evaluated, filtered, and expanded through controlled error insertion techniques. These methods allowed us to create a large-scale dataset with diverse translation quality scores, which was subsequently used to train neural QE models. Our experiments highlight the impact of dataset size, distribution balance, and targeted linguistic issues—such as grammatical gender and number agreement errors—on model performance.

This paper describes the challenges encountered in training QE models on English-Hebrew segments, our methodology for constructing the dataset, and the findings from our experiments. We also discuss future directions aimed at improving QE performance by incorporating additional linguistic phenomena, such as handling idiomatic expressions.

## 2. Building the Dataset

Quality estimation for English-Hebrew translation presents distinct challenges stemming from the linguistic differences between the two languages. Hebrew is a morphologically rich language, exhibiting extensive inflection for gender, number, and tense, which often lack explicit markers in English source texts. This discrepancy leads to frequent translation errors, particularly in subject-verb agreement and noun-adjective concord. Moreover, Hebrew's flexible word order and reliance on contextual inference make it difficult for MT systems—and consequently for QE models—to consistently capture grammaticality and naturalness. These factors are further compounded by the relatively limited availability of large, high-quality parallel corpora. As a result, developing robust QE models for this language pair requires careful dataset construction to expose models to a wide range of linguistic phenomena, controlled error types, and varied translation quality levels.

To address these challenges, we constructed a semi-synthetic parallel dataset designed to enhance the reliability of QE models. Synthetic parallel corpora have been widely used to improve translation quality in under-resourced languages (Edunov et al. 2018). However, while synthetic data generation is common in QE, ensuring that the dataset accurately represents diverse translation quality remains an ongoing research challenge.

First, we constructed a dataset of examples of usage extracted from an intermediate-level English learner's dictionary. These examples stem from typical linguistic patterns that were identified by experts in reliable corpora and thus served as a base for compiling full sentences and short phrases that illustrate typical usage for language learners. This makes these examples of usage particularly suitable foundational references for generating new sentences automatically, using ChatGPT. Specifically, ChatGPT 3.5 was prompted to generate sentences that align with the linguistic structure and semantic content of each original usage example (cf. Appendix A for a prompt example).

The generated sentences were then translated into Hebrew using three MT engines: Google Translate, Yandex Translator, and Microsoft Translator. To assess translation similarity, we computed the BLEU score between each pair of translations. Sentences whose translations did not achieve a BLEU score of at least 0.85 were excluded from the dataset. Among the remaining segments, we selected the translation pair with the highest agreement score, under the assumption that consensus between two different engines indicates higher translation quality. Similar filtering strategies have been explored in QE research to ensure a balanced dataset (Ranasinghe et al. 2020). However, a post hoc analysis revealed that many segments exhibited systematic translation errors common across the MT engines.

## 3. Translation Ranking

The quality of our English-Hebrew parallel corpus was manually assessed to ensure its reliability and usability. This evaluation focused on multiple aspects, including accuracy, linguistic quality, style, readability, and overall functionality (cf. Appendix B for the ranking guidelines).

The translations were ranked on a five-point scale, ranging from excellent translations requiring no corrections to extremely poor translations deemed unacceptable. Errors were classified based on their severity, with distinctions made between neutral stylistic differences, minor inaccuracies affecting precision without significantly altering meaning, and major errors that critically impacted interpretation.

Accuracy was a key criterion, emphasizing content fidelity and consistent terminology use, particularly in specialized domains such as legal, medical, and technical texts. The evaluation also assessed linguistic quality, ensuring adherence to grammatical and syntactic norms, proper punctuation, and correct spelling. Instances of inappropriate punctuation were considered as minor errors, while more significant grammatical or syntactic inconsistencies were classified as major.

The assessment extended to stylistic and tonal fidelity, requiring translations to preserve the register, formality, and tone of the source text while also being culturally appropriate. Translations exhibiting non-versatile vocabulary or excessive repetition were penalized, particularly when that affected clarity and natural flow. Readability and fluency were also integral to the evaluation process, with preference given to translations that maintained a natural linguistic structure over overly literal interpretation. Ensuring completeness was another factor, where all content from the source text was expected to be included unless omissions were explicitly required for clarity.

Functional equivalence was essential in evaluating translations of instructional, legal, and marketing texts, where the purpose had to be effectively conveyed in the translation. Machine-generated translations that adhered too closely to a literal structure often led to awkward phrasing, affecting coherence and clarity. Additionally, the quality of synthetic source texts themselves was evaluated, with English sentences deemed illogical or containing irrelevant characters automatically receiving lower scores, regardless of the translation quality.

By implementing this structured manual evaluation framework, approximately 14,000 ranked segments were produced, maintaining high linguistic and functional standards, to ensure the corpus's suitability for further research and practical applications in translation studies and MT development.

## 4. Perturbation-Based Dataset Augmentation

We trained BERT (Devlin et al. 2019), XLM-R (Conneau et al. 2020), and TransQuest (Ranasinghe et al. 2020) on this dataset to perform the QE task. A fully connected (FC) neural network was integrated as a classification head on top of the pre-trained BERT and XLM-RoBERTa (XLM-R) models, enabling the transformation of model outputs into quality scores. This classification head comprises four layers and processes the final hidden state, which is sequentially passed through fully connected layers of decreasing dimensions (512 → 256 → 64), ultimately generating a score prediction within five categories ranging from 1 to 5. However, the results were unsatisfactory, leading to the conclusion that the dataset was too small to serve as a reliable resource for training a model capable of learning the complexities of translation quality assessment between English and Hebrew. This was particularly evident in the skewed distribution of quality scores, with the vast majority of segments receiving scores of 3 and 5.

Our attempts to train the TransQuest model (both Mono and Siamese versions using BERT Base and XLM-R) failed to achieve a Pearson correlation above 0.1. Additionally, training Bert Base for sequence classification and XLM-R resulted in models that predominantly assigned a score of 3 to test dataset.

To address these limitations, we introduced controlled translation errors into the dataset, modifying quality scores accordingly based on the severity of errors. Such controlled data augmentation techniques have been successfully used in previous research to improve morphological and syntactic agreement in MT systems (Habash and Roth 2009; More and Tsarfaty 2016).

## 4.1 Grammatical error-based augmentation

In Hebrew, verb inflections and morphosyntactic agreement play a crucial role in sentence correctness. Alterations in verb inflections for gender (masculine vs. feminine) and number (singular vs. plural) can lead to syntactic anomalies and semantic distortions. These phenomena stem from the language's rich agreement system, which requires agreement between verbs and their subjects, nouns and their modifying adjectives, and, in some cases, additional sentence elements. Such errors are prevalent among Hebrew language learners and pose significant challenges in natural language processing (NLP) applications, where statistical models may generate grammatically incorrect structures.

In the context of NLP, these challenges manifest in tasks such as part-of-speech tagging and dependency parsing, where recognizing and correctly classifying verb inflections is crucial. In MT, errors in gender and number agreement can lead to incorrect translations, particularly when the source sentence does not explicitly indicate agreement constraints. Neural models relying on computational representations often require a broader context to mitigate such errors, highlighting the complexity of morphosyntactic agreement in morphologically rich languages like Hebrew.

To address these challenges, we utilized expert-developed morphological lexicons that classify words according to grammatical features such as number and gender, clearly differentiating between singular and plural forms, as well as masculine and feminine variants. We used these lexicons to systematically expand the original dataset by introducing controlled, artificial errors.

For each segment, the introduction of a single error resulted in a one-point reduction in its translation quality score, while two errors led to a two-point reduction. Given the limited availability of segments rated with a quality score of 4, we supplemented the data with two professionally translated English-Hebrew datasets, automatically assigning them a quality score of 5. The same error-insertion methodology was applied to these datasets, resulting in most sentences being rated at quality levels 4 and 3. Through this process, we constructed a parallel dataset of approximately 200,000 segments, each annotated with a corresponding quality score.

## 4.2 Adversarial negatives augmentation

With the expanded dataset, we trained BERT Base model for sequence classification, which produced more reliable quality estimates, ultimately achieving a Pearson correlation of 0.85 on the test set.

However, manual evaluations revealed a notable limitation: while the model performed well when evaluating Hebrew translations of the corresponding English sentences, it produced unreliable scores when presented with segments in which the Hebrew sentence was randomly chosen and did not correspond to the English source. In such cases, instead of consistently assigning a quality score of 1, the model exhibited varied predictions.

To address this issue, we further doubled the dataset from 200,000 to 400,000 segments by pairing English sentences with randomly selected, unrelated Hebrew sentences. These mismatched sentence pairs were explicitly assigned a translation quality score of 0. Using this extended dataset, we retrained the BERT model, but this time the model struggled to surpass a Pearson correlation of 0.2. Moreover, it exhibited a strong bias toward assigning a score of 0 to test segments, even when the translation quality of those segments was objectively high. To counteract this issue, we reduced the proportion of segments with a score of 0 to one-third of the training dataset. The final dataset thus consisted of 200,000 translated and reliably scored segments and 100,000 segments where an English sentence was paired with an unrelated Hebrew sentence, explicitly marked with a quality score of 0.

### 4.3 Word order augmentation

To further enrich the dataset and improve model generalization, we applied a series of perturbation-based data augmentation techniques that simulate syntactic and semantic distortions. For each English-Hebrew segment, we introduced variations by modifying the word order in the Hebrew translation. In the first variation, each word was swapped with its adjacent word, and the resulting segment was assigned a quality score reduced by one point. In the second variation, each word was shifted two positions forward, and the corresponding segment received a two-point penalty due to its increased structural distortion. In cases where fewer than 20 perturbed segments were generated through these methods, we supplemented the set by randomly shuffling the words in the translation to complete a batch of 20 augmented variants. These randomly shuffled segments, exhibiting more severe syntactic disintegration, were penalized by three points. Such perturbation strategies have been shown in prior research to effectively enhance the robustness of language models by exposing them to controlled noise and structural variations (Belinkov and Bisk 2018).

### 5. Experiments

To evaluate the effectiveness of our approach, we conducted a series of experiments using a modified version of the pre-trained BERT base model for sequence classification. Model training was performed using the LoRA technique (Hu et al. 2022), configured with a rank of 8, an alpha value of 16, and a dropout rate of 0.1 to improve training efficiency and reduce overfitting.

We trained the model on multiple datasets that varied in sampling strategy and corpus size, aiming to assess how differences in data distribution and volume influence the model's performance in the translation quality estimation task.

### 5.1 Normal distribution sampling

To further assess the impact of dataset size and distribution on model performance, we constructed a dataset of 500,000 segments by randomly sampling from our larger pool of parallel English-Hebrew sentence pairs. The sampling process ensured a relatively balanced representation across the range of quality scores, producing a distribution that approximates a normal curve.

When trained on the normally distributed dataset, the BERT-based QE model attained a Pearson correlation of 0.65 on the test set, reflecting a moderate agreement between predicted and actual quality scores. This outcome establishes a baseline for evaluating the effectiveness of alternative data selection strategies presented in the following sections.

### 5.2 Uniform distribution sampling

To investigate the effect of score distribution on model performance, we constructed a second dataset comprising 430,000 English-Hebrew sentence pairs. Unlike the previously discussed dataset, which followed a normal distribution, this dataset was explicitly sampled to reflect a uniform distribution across all six quality score levels (ranging from 0 to 5). Each score level was equally represented, ensuring that the model would encounter a balanced number of examples for each quality category during training.

This uniform distribution was designed to mitigate the skew typically observed in naturally occurring translation corpora, where higher quality translations (i.e., scores of 4 and 5) often dominate the dataset. By exposing the model equally to all levels of translation quality, we aimed to encourage more nuanced learning, particularly in distinguishing between mid-range and low-quality translations, which are often under-represented in standard datasets.

Training the BERT-based quality estimation model on this uniformly distributed dataset resulted in a Pearson correlation coefficient of 0.88 on the test set—a substantial improvement over the 0.65 achieved with the normally distributed dataset. This result suggests that balancing the training data across all quality levels enables the model to learn a more fine-grained and consistent mapping from input segments to quality scores.

Our findings underscore the importance of considering score distribution when constructing datasets for QE. A uniform sampling strategy not only enhances the model's predictive accuracy but also reduces the default tendency to the mean score, a common failure mode observed in imbalanced training scenarios. This experiment demonstrates that strategic dataset design can significantly influence model performance and provides motivation for further exploration of distribution-aware sampling techniques.

## 5.3 Large-scale random sampling

To further examine the influence of dataset scale on model performance, we conducted additional experiments using larger corpora consisting of 1 million and 4 million randomly sampled English-Hebrew segments. The sampling process preserved the overall characteristics of the source corpus without explicitly enforcing a target distribution, allowing us to observe how volume alone, combined with natural variability, would impact model training.

The results were significant. Both datasets yielded the highest performance observed across all experiments, with the model trained on each achieving a Pearson correlation of 0.92 on the held-out test set. This strong correlation indicates a highly accurate alignment between predicted and reference quality scores and demonstrates the model's improved capacity to discriminate between fine-grained levels of translation quality.

These findings underscore two critical aspects of QE model development: the importance of training data scale and the underlying distribution of quality scores. While balanced distributions (as discussed in Section 5.2) improve performance by reducing prediction bias, the results from this experiment show that scale can independently enhance learning effectiveness. Having access to a broader spectrum of linguistic variation—both syntactic and semantic—the model benefits from richer training signals and reduced overfitting, ultimately leading to more generalizable performance.

Moreover, our findings align with prior work emphasizing the value of large-scale, perturbed, or semi-synthetic datasets in under-resourced MT contexts (Fadaee et al. 2017). These results reinforce the notion that carefully constructed large datasets, even if partially composed of noisy or artificially modified examples, can significantly boost the robustness and sensitivity of QE models in real-world translation pipelines. This suggests that data augmentation and scaling strategies should be jointly considered in future efforts to improve quality estimation, especially for morphologically rich or under-resourced language pairs.

## 6. Future Work

In future research, we aim to further refine the model's ability to assess translation quality by addressing a common issue observed in commercial translation engines: the inaccurate translation of English idioms into Hebrew, which remain a major challenge in MT. We have compiled a list of English idioms and plan to augment our dataset with segments containing idioms alongside their correct Hebrew equivalents, as well as segments featuring erroneous translations of these idioms. We anticipate that incorporating this data will enhance the model's performance in evaluating the quality of English-to-Hebrew translations.

**Appendix A – Sentence Generation Prompt**

Below is an example of the prompt used to generate sentences based on usage examples taken from *Password* English learner's dictionary.

Prompt: Taken from high-school English learner's dictionary – the dictionary entry of the headword: "carnation", part-of-speech: "noun", has the following example sentence: "All the men wore carnations in their buttonholes." – suggest an additional sentence that contains at least 20 words and that corresponds to the existing example sentence in terms of linguistic structure and academic level.

Result: The little girl handed her teacher a handmade card with a delicately pressed carnation, expressing her gratitude and admiration.

**Appendix B – Translation Ranking Guidelines**

The purpose of these guidelines is to provide a framework for evaluating the quality of machine translations. The evaluation criteria are designed to assess not only the accuracy of the translation but also its linguistic quality, style, readability, and overall functionality.

**Ranking Scale**

The translation quality is assessed on a 1 to 5 scale, defined as follows:

- 5 – excellent translation; no corrections needed
- 4 – good translation; minor improvement(s) need to be made
- 3 – medium translation; major improvement(s) need to be made
- 2 – poor translation; many improvements need to be made
- 1 – extremely poor translation

**Error Categorization**

Errors detected in the evaluation process are categorized into three levels based on their severity and in regard to the criteria that is discussed below:

- **Neutral (fine):** acceptable translations with a minor modification to be done due to either stylistic differences or personal preferences, but not constituting crucial errors.
- **Minor (inaccurate):** errors that affect the precision of the translation but do not significantly alter the meaning; translations contain minor grammatical errors or slight terminology inconsistencies.
- **Major (problematic):** errors that are critical and impact the accuracy or interpretation of the translation; these errors include incorrect translations of key concepts, poor grammar, or terminological mistakes in specialized fields.

**Evaluation Criteria**

The quality of translations is to be determined based on the following:

1. **Accuracy:**

**Content Fidelity:** the translation is to accurately reflect the meaning of the original text; this includes preserving the nuances, context, and intent of the source material.

**Terminology Consistency:** the correct and consistent use of terminology is crucial, especially in specialized texts (e.g., legal, medical, technical).

An example for an accuracy-inflected problem is a fine translation in Hebrew that demonstrates inconsistencies in the grammatical number of persons; such translation receives a 3 score:

(a) *The lifeguard quickly responded to the swimmer in distress and brought them back to shore, ensuring they were safe and sound.*

המציל הגיב במהירות לשחיין במצוקה והחזיר אותם לחוף, וידא שהם בריאים ושלמים.

The ambiguous gender of the swimmer in the English sentence resulted in a misinterpretation of the automated translation, having the swimmer in the singular masculine form, and the other pronouns in the plural form; that error would count as **major**.

2. **Linguistic Quality:**

**Grammar and Syntax:** the translation should adhere to the grammatical rules of the target language; i.e., proper syntactical structure, verb conjugation, and correct usage of articles and prepositions.

**Punctuation and Spelling:** proper punctuation and spelling are essential in maintaining clarity and coherence in the translations.

Inconsistencies such as those illustrated in sentence (a) are acceptable in Hebrew and receive a 4 score in some cases of general sayings that don't indicate a specific recipient. For example, the translation in (b) presents twice the use of the singular form, and once the plural form, thus counting as a **minor** error:

> **(b)** It's important not to be too hard on yourself when you make a mistake; remember that everyone has room to grow and learn.
>
> חשוב לא להיות קשה מדי עם עצמך כאשר אתה עושה טעות; זכרו שלכל אחד יש מקום לצמוח וללמוד.

Missing punctuation marks count as a **minor** error and result in a 4 score. For example:

> **(c)** The spicy salsa gave me a burning sensation in my mouth, making me reach for a glass of cold water.
>
> הסלסה החריפה נתנה לי תחושת צריבה בפה וגרמה לי להושיט יד לכוס מים קרים

Irrelevant punctuation marks that don't appear in the original sentence count as a **minor** error result in a 4 score. For example:

> **(d)** The school board decided to transition from a single-gender institution to a co-educational one, in order to provide equal opportunities for all students.
>
> "הנהלת בית הספר החליטה לעבור ממוסד חד-מגדרי למוסד חינוכי משותף, על מנת לספק שוויון הזדמנויות לכל התלמידים."/

Grammatical and/or syntactic rules that are wrongly applied or are not applied at all in the conversion from English to the target language receive a 3 score; the MT tool might follow the English punctuation and grammar, even when these are incompatible in the target language. However, this point is exceptional in cases the editor finds it to be appropriate, based on context, counting as a **minor** error and resulting in a 4 score. For example:

> **(e)** Our new neighbor is extremely approachable, and we feel comfortable asking her for guidance on local activities and places to visit.
>
> השכנה החדשה שלנו נגישה מאוד ,ואנחנו מרגישים בנוח לבקש ממנה הדרכה על פעילויות מקומיות ומקומות לבקר בהם.

When having a conjunction/connector in Hebrew (i.e., Vav HaChibur), no comma appears before it. However, when having a clause or a conjunction in English, there is a comma before the last component; the MT tool places commas inappropriately in the Hebrew translation. Even though the comma in the segment above is unnecessary, it indicates a separation of two clauses, and therefore was excusable.

3. **Style and Tone:**

**Stylistic Fidelity:** the style of the translation should match that of the original text in register, tone, formality, etc.

**Cultural Appropriateness:** the translation should be culturally sensitive and adapted to the norms and expectations of the target audience and with accordance to the type of text in the source language. Translations that are correct but have non-versatile vocabulary are given a 3 score. For example:

(f) *She became a pop star overnight after her debut single topped the charts, captivating audiences with her mesmerizing vocals and unique style.*

היא הפכה לכוכבת פופ בן לילה לאחר שסינגל הבכורה שלה כבש את ראש המצעדים, וכבש את הקהל עם השירה המהפנטת והסגנון הייחודי שלה.

The two Hebrew terms marked in green are identical and would be translated back to English as 'conquered'. Moreover, the second term is of a wrong inflectional form, making it seem like the debut single is the one to captivate audiences, rather than the singer herself. Lastly, the terms marked in blue use the possessive preposition unnecessarily, resulting in repetitive, non-versatile vocabulary; all these errors count as a **major** severity level.

The Hebrew translation could be improved by, e.g., applying the following modifications:

*היא הפכה לכוכבת פופ בן לילה לאחר שסינגל הבכורה שלה כבש את ראש המצעדים, ושבתה את הקהל עם שירתה המהפנטת וסגנונה הייחודי.*

A 4 score is given to sentences that are translated correctly but can be improved. In sentence (g) below, the Hebrew translation is correct and valid in colloquial language; nevertheless, it suffers from a **minor** error, which is the absence of the interrogative word האם:

(g) *Did you enjoy reading the funny stories depicted in the comic strips featured in yesterday's newspaper?*

*נהנית לקרוא את הסיפורים המצחיקים המתוארים ברצועות הקומיקס שהופיעו בעיתון של אתמול?*

4. **Readability and Fluency:**

**Natural Flow:** the translated text should read naturally in the target language, without awkward phrasing or unnatural constructions.

**Clarity:** the translation should be clear and easy to understand, avoiding ambiguity or confusion.

When sentences are translated literally rather than figuratively; and/or have a wrong choice of words, a 2-3 score is given, as these errors also count as **major**. For example:

(h) *The team is determined to get their new business idea off the ground by securing investors and developing a solid marketing strategy.*

*הצוות נחוש להוציא את הרעיון העסקי החדש שלהם מהקרקע על ידי אבטחת משקיעים ופיתוח אסטרטגיה שיווקית מוצקה.*

An improved, non-literal translation conveying the figurative meaning illustrated in the English segment could be:

*הצוות נחוש למנף את הרעיון העסקי החדש שלו על ידי אבטחת משקיעים ופיתוח אסטרטגיה שיווקית מוצקה.*

5. **Completeness:**

**Full Translation:** the entire content of the source text should be translated unless specified otherwise; omissions or additions should only occur when they are necessary for clarity or when they are explicitly requested.

6. **Functional Equivalence:**

**Purpose Alignment:** the translation should align with the function of the source text; this is particularly important in instructional materials, legal documents, and marketing content.

**Literal vs. Nonliteral Translations**: automated tools might generate overly literal translations that follow the source language's word-for-word structure, leading to awkward or incorrect results in the target language; the editor is to assess whether the translation captures the intended meaning in a natural-like language. Unnaturalness counts as a **major** error, and translations that seem unnatural and/or irrational receive a 2-3 score, depending on severity. For example:.

(i) *Our teacher assigned us a challenging text from a classic novel for our next literature discussion.*

<div dir="rtl">*המורה שלנו הקצה לנו טקסט מאתגר מרומן קלאסי <u>לדיון הספרות</u> הבא שלנו.*</div>

The Hebrew sentence is correct when translated literally, but the marked terms seem unnatural and even incorrect in that context; it would've been better translated into:

<div dir="rtl">*המורה שלנו הקצה לנו טקסט מאתגר מרומן קלאסי <u>לדיון הספרותי</u> הבא שלנו.*</div>

Or rephrased as:

<div dir="rtl">*המורה שלנו הקצה לנו טקסט מאתגר מרומן קלאסי <u>לדיון הבא בשיעור ספרות</u>.*</div>

**Additional rules for evaluating the quality of source texts:**

English sentences that are unnatural and/or irrational receive a <u>1 score</u>, and their translations are not reviewed. For example:

**(j)** *The action of the car engine requires regular maintenance to ensure optimal performance.*

Sentence (j) is illogical since it regards the maintenance of 'the action of', while it should refer to the engine itself.

Some English sentences having strange and irrelevant characters, result in a <u>3 score</u> regardless of their translation. For example:

**(k)** *If you take the square root of 64, you'll get 8 (â64 = 8), which tells us that 8 multiplied by itself equals 64.*

## 6. References

Belinkov, Y., and Y. Bisk. 2018. Synthetic and natural noise both break neural machine translation. *Proceedings of the 6th International Conference on Learning Representations (ICLR 2018)*.

Conneau, A., Khandelwal, K., Goyal, N., Chaudhary, V., Wenzek, G., Guzmán, F., Grave, E., Ott, M., Zettlemoyer, L., and V. Stoyanov. 2020. Unsupervised Cross-lingual Representation Learning at Scale. In *Proceedings of the 58th Annual Meeting of the Association for Computational Linguistics*, 8440–8451, Online. Association for Computational Linguistics. https://aclanthology.org/2020.acl-main.747/

Devlin, J., Chang, M. W., Lee, K., and K. Toutanova. 2019. BERT: Pre-training of deep bidirectional transformers for language understanding. *Proceedings of the 2019 Conference of the North American Chapter of the Association for Computational Linguistics* (NAACL-HLT 2019).

Edunov, S., Ott, M., Auli, M., and D. Grangier. 2018. Understanding back-translation at scale. *Proceedings of the 2018 Conference on Empirical Methods in Natural Language Processing (EMNLP)*, 489–500.

Fadaee, M., Bisazza, A., and C. Monz. 2017. Data augmentation for low-resource neural machine translation. *Proceedings of the 55th Annual Meeting of the Association for Computational Linguistics* (ACL 2017), 567–573. Vancouver, Canada. Association for Computational Linguistics.

Habash, N., and R. Roth. 2009. CATiB: The Columbia Arabic Treebank. *Proceedings of the Workshop on Computational Approaches to Semitic Languages*, 1-8.

Hu, E., Shen, Y., Wallis, P., Allen-Zhu, Z., Li, Y., Wang, L., and W. Chen. 2022. LoRA: Low-Rank Adaptation of Large Language Models. arXiv preprint arXiv:2106.09685.

Ive, I., Blain, F., and L. Specia. 2018. deepQuest: A framework for neural-based quality estimation. In *Proceedings of the 27th International Conference on Computational Linguistics, 3146–3157, Santa Fe, New Mexico, USA, August* 2018. Association for Computational Linguistics.

Kepler F., Trenous, J., Treviso, M., Vera, M., and A. F. T. Martins. 2019. OpenKiwi: An open source framework for quality estimation. In *Proceedings of the 57th Annual Meeting of the Association for Computational Linguistics: System Demonstrations, 117–122, Florence, Italy, July 2019*. Association for Computational Linguistics. Hyun Kim, Jong-Hyeok Lee, and S

More, A., and R. Tsarfaty. 2016. Data-driven morphological analysis and disambiguation for morphologically rich languages (MRLs). *Transactions of the Association for Computational Linguistics* (TACL), 4, 1-20.

Ranasinghe, T., Orasan, C., and R. Mitkov. 2020. TransQuest: Translation quality estimation with cross-lingual transformers. *Proceedings of the 28th International Conference on Computational Linguistics* (COLING 2020).

Specia, L., Blain, F., Logacheva, V., Astudillo, R. and A. F. T. Martins. 2018. Findings of the WMT 2018 shared task on quality estimation. In *Proceedings of the Third Conference on Machine Translation: Shared Task Papers, 689–709, Belgium, Brussels, October 2018*. Association for Computational Linguistics.